\documentclass{article}
\usepackage[preprint]{spconf}
\usepackage{amsmath,graphicx}
\usepackage{algpseudocode}
\usepackage{algorithm} 
\usepackage{booktabs}
\usepackage{cite}
\usepackage{placeins} 
\usepackage{url}
\usepackage[bottom]{footmisc}

\newcommand{\cat}{\text{{Cat}}}

\title{SELF-SUPERVISED REPRESENTATIONS IN SPEECH-BASED DEPRESSION DETECTION}
\name{Wen Wu\thanks{Wen Wu is supported by a Cambridge International Scholarship from the Cambridge Trust. This work has been performed using resources provided by the Cambridge Tier-2 system operated by the University of Cambridge Research Computing Service (www.hpc.cam.ac.uk) funded by EPSRC Tier-2 capital grant EP/T022159/1. 
The MSP-Podcast data was provided by The University of Texas at Dallas through the Multimodal Signal Processing Lab. This material is based upon work supported by the National Science Foundation under Grants No. IIS-1453781 and CNS-1823166. Any opinions, findings, and conclusions or recommendations expressed in this material are those of the author(s) and do not necessarily reflect the views of the National Science Foundation or The University of Texas at Dallas.
}, Chao Zhang, Philip C. Woodland}
\address{Department of Engineering, University of Cambridge, Trumpington St., Cambridge, UK.\\
\small{\texttt{\{ww368,cz277,pcw\}@eng.cam.ac.uk}}}
\copyrightnotice{\copyright\ IEEE 2023}
                \toappear{To appear in {\it Proc.\ ICASSP2023,
                   June 04-10, 2023, Rhodes, Greece}}

\begin{document}
\ninept
\maketitle

\begin{abstract}
This paper proposes handling training data sparsity in speech-based automatic depression detection (SDD) using foundation models pre-trained with self-supervised learning (SSL). An analysis of SSL representations derived from different layers of pre-trained foundation models is first presented for SDD, which provides insight to suitable indicator for depression detection. Knowledge transfer is then performed from automatic speech recognition (ASR) and emotion recognition to SDD by fine-tuning the foundation models. 
Results show that the uses of oracle and ASR transcriptions yield similar SDD performance when the hidden representations of the ASR model is incorporated along with the ASR textual information. By integrating representations from multiple foundation models, state-of-the-art SDD results based on real ASR were achieved on the DAIC-WOZ dataset.
\end{abstract}
\section{Introduction}
\label{sec:intro}

Depression is a serious mood disorder affecting about 280 million people in the world~\cite{WHO}, and at present there is no objective measure for depression detection with clinical utility~\cite{cummins2015review}. In order to develop a fully automatic depression detection system, a growing body of research has demonstrated that
correlations of depression are detectable in spontaneous speech~\cite{moore2007critical,low2010detection,ooi2012multichannel,williamson2016detecting, al2018detecting}. Despite encouraging progress, speech-based depression detection (SDD) is still challenging due to the variability in depression manifestations and lack of training data.

Foundation models refer to single universal models trained on broad data at scale that can be used in a variety of related downstream tasks and domains~\cite{bommasani2021opportunities}. Recently, foundation models have sparked a research paradigm shift in many fields of artificial intelligence. 
Self-supervised learning (SSL) is a prevalent approach to pre-train a foundation model, in which the training labels are extracted from the input features themselves thus enabling the use of a large amount of unlabelled training data.
It has been shown that SSL representations, the intermediate layer output of an SSL pre-trained foundation model, are often useful for many downstream tasks~\cite{yang2021superb}. In particular, speech foundation models, such as wav2vec 2.0 (W2V2)~\cite{baevski2020wav2vec}, HuBERT~\cite{hsu2021hubert}, and WavLM~\cite{chen2022wavlm}, are attracting increasing attention and have achieved state-of-the-art (SOTA) results in many speech processing tasks, including automatic speech recognition (ASR) and automatic emotion recognition (AER)~\cite{zhang2022bigssl,morais2022speech}, \textit{etc}. Despite this great success, SSL representations have not been extensively studied for SDD.

This paper studies the use of SSL-pretrained speech foundation models to handle the challenges in SDD. This allows the data sparsity issue to be handled via large amount of unlabelled data used for SSL pre-training. Such unlabelled data can be produced by many speakers that cover much speaker variability and hence can help to model speaker-dependent depression manifestation variability.   
A block-wise analysis was first performed to compare the SSL representations from different layers of different foundation models and to understand what type of information is more effective in SDD. Next, the foundation models were fine-tuned for ASR and AER tasks separately, to investigate the knowledge transfer from ASR and AER to SDD and the effect of fine-tuning on the intermediate layers. Three different speech foundation models, W2V2, HuBERT and WavLM, were compared. ASR transcriptions were encoded by RoBERTa~\cite{liu2019roberta}, a text foundation model, and incorporated. The ensemble with multiple foundation models gives SOTA results on the benchmark DAIC-WOZ dataset~\cite{DAIC-WOZ}.

The rest of the paper is organised as follows. Section~\ref{sec: method} introduces the proposed method and the experimental setup. Sections~\ref{sec: speech ssl} and \ref{sec: text ssl} present the block-wise analysis of speech foundation models and the use of ASR transcriptions in depression detection respectively. The foundation models are combined in Section~\ref{sec:combination}, followed by conclusions.

\section{PROPOSED MODEL}
\label{sec: method}
\subsection{Model structure}

\begin{figure}[tb]
    \centering
    \includegraphics[width=0.9\linewidth]{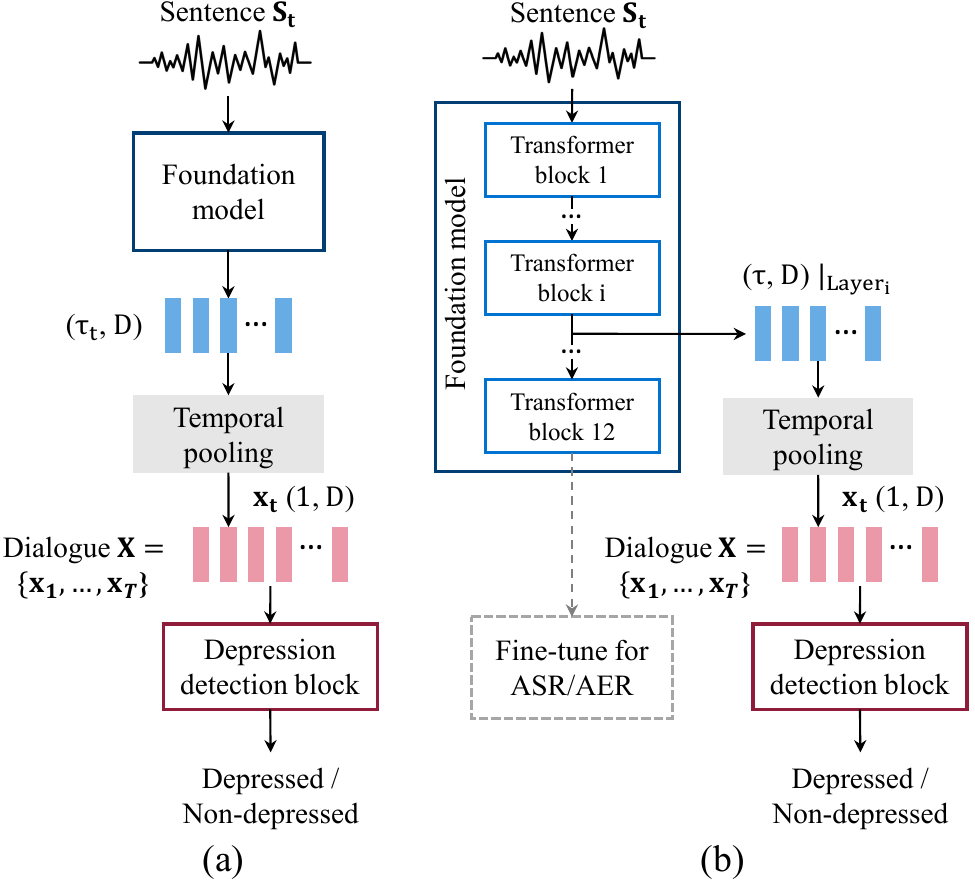}
    \vspace{-2ex}
    \caption{(a) Model structure. (b) The block-wise analysis framework.}
    \label{fig: updown}
    \vspace{-2ex}
\end{figure}

In this paper, SDD is formulated as a binary classification task that determines whether the speaker is depressed or not. The model structure is illustrated in Fig.~\ref{fig: updown}(a) which contains a foundation model followed by a depression detection block. The SDD system takes a dialogue $\mathbf{X}$ (\textit{i.e.} a clinical interview) as input, which consists of a sequence of sentences $\mathbf{X}=\{\mathbf{x}_1, ... , \mathbf{x}_T\}$ where $T$ is the number of utterances in the dialogue. The foundation model takes an utterance $\mathbf{S}_t$ as the input and produces a vector of size $(\tau_t, D)$ where $\tau_t$ is the number of frames in $\mathbf{x}_t$ and $D$ is the feature dimension.  Temporal pooling (average pooling was used in this paper) is then applied to the output of the foundation model, producing a $D$ dimensional (-dim) vector for each utterance. The depression detection block then takes a dialogue consisting of $T$-length sequence with $D$-dim vectors as its to perform the diagnosis.

Three pre-trained foundation models were used in this paper: wav2vec 2.0\footnote{Available at https://huggingface.co/facebook/wav2vec2-base} (W2V2), HuBERT\footnote{Available at https://huggingface.co/facebook/hubert-base-ls960}, and WavLM\footnote{Available at https://huggingface.co/microsoft/wavlm-base-plus}. The base versions were used for all three foundation models which contain twelve 768-dim Transformer encoder blocks and about 95M parameters.  The depression detection block consists of two 128-dim Transformer encoder blocks with four attention heads each, followed by a fully-connected (FC) output layer. Transformer structure was chosen as it's the de facto standard model in sequence modelling tasks. The depression detection block has 0.3M parameters.

\subsection{Dataset}
DAIC-WOZ~\cite{DAIC-WOZ}, a benchmark dataset for depression detection, consists of 189 clinical interviews between an interviewer and a patient. 30 out of 107 interviews within the training set and 12 out of 35 interviews within the development set are classified as depressed. Classification performance is evaluated by the F1 score. Following prior work~\cite{shen2022automatic,gong2017topic,ravi22_interspeech,al2018detecting,wu2022climate}, results on the development subset are reported. The model was initialised and trained for 20 different random seeds and both the highest (F1-max) and the average (F1-avg) value are reported, along with the standard deviation (F1-std) across seeds. 

\begin{algorithm}[tb]
	\begin{minipage}{\linewidth}
	\caption{Sub-dialogue shuffling} 
	\begin{algorithmic}[1]
	    \State $N^{+} \leftarrow$ Number of positive samples in the training set
	    \State $N^{-} \leftarrow$ Number of negative samples in the training set
	    \State Set number of sub-dialogues for each positive sample $M^{+}$
	    \State $M^{-}\leftarrow{N^{+}\times M^{+}}/{N^{-}}$
	    \State Set $\epsilon_l$, $\epsilon_h$ satisfying $0<\epsilon_l<\epsilon_h<=1$
		\For {Dialogue $\mathbf{X}^{(n)}, n=1,2,\ldots,N$}
    		\State $T\leftarrow\operatorname{len}(\mathbf{X}^{(n)})$
    		\If {$\mathbf{X}^{(n)}$ is positive} $\  M \leftarrow M^{+}$ 
    		\Else $\quad M \leftarrow M^{-}$
    		\EndIf
			\For {Sub-dialogue $\mathbf{X}^{(n)_m}, m=1,2,\ldots,M$}
			    \State Sample $\epsilon$ uniformly from $[\epsilon_l,\epsilon_h)$
			    \State $d \leftarrow \epsilon T -1$
				\State Sample $s$ randomly from range $[0,T-d)$
				\State $e \leftarrow s+d$
				\State $\mathbf{X}^{(n)_m} \leftarrow \mathbf{X}^{(n)}_{s:e}$
			\EndFor
		\EndFor
	\end{algorithmic}
	\label{al: data aug}
	\end{minipage}
\end{algorithm}

\subsection{Data augmentation}
Depression is usually assessed by clinical interview and labelled at the session-level, which results in one label per interview. Given a certain amount of data, the number of samples in an SDD dataset is usually much smaller than the number of utterances and frames often used in other speech tasks (\textit{e.g.} speech and speaker recognition), which makes SDD 
a very data sparse scenario. 
For instance, DAIC-WOZ consists of 50+ hours of speech recordings that correspond to merely 189 samples. Privacy concerns and labelling difficulty further increase the data sparsity issue in SDD. 
Furthermore, data imbalance is another severe issue since the positive cases are much fewer than negative cases (28\% vs. 72\% in training). Therefore, it is crucial to use data augmentation to alleviate both data scarcity and imbalance issues for SDD.

In this paper, the training set was augmented using sub-dialogue shuffling, which samples a sub-dialogue $\mathbf{x}_{s:e}$ from each complete dialogue $\mathbf{x}_{1:T}$, where $s$ and $e$ are the randomly selected start and end utterance indexes. The details are given in Algorithm~\ref{al: data aug}. Firstly, the number of positive and negative samples in the training set are counted and $M^{+}$ is set which is the desired number of sub-dialogues for each positive dialogue (line 1-3 of Algorithm~\ref{al: data aug}). To augment while balancing the training samples, $M^{-}$ is computed based on $N^+, N^-$, and $M^+$ (line 4). Then, $M^{+}$ and $M^{-}$ sub-dialogues are generated for each complete dialogue belonging to the positive and negative classes respectively (line 8-10 of Algorithm~\ref{al: data aug}). $\epsilon_l$ and $\epsilon_h$ are two variables that determine the length range of the sub-dialogues. When generating a sub-dialogue, its length $d$ is first defined by a coefficient randomly drawn from $[\epsilon_l,\epsilon_h)$ (line 12-13). The start index $s$ is then randomly chosen from its available range and the end index is then determined (line 14-16).

\begin{table*}[tb]
\centering
\begin{tabular}{cccc|cccc|cccc}
\toprule
\multicolumn{4}{c|}{W2V2$^\text{PT}$}       & \multicolumn{4}{c|}{HuBERT$^\text{PT}$}     & \multicolumn{4}{c}{WavLM$^\text{PT}$}      \\
\midrule
Block & F1-avg & F1-max & F1-std & Block & F1-avg & F1-max & F1-std & Block & F1-avg & F1-max & F1-std \\
2     & 0.531  & 0.615  & 0.044  & 2     & 0.557 & 0.615  & 0.033  & 2     & 0.545  & 0.636  & 0.033  \\
4     & 0.549  & 0.667  & 0.055  & 4     & 0.582 & 0.621  & 0.020  & 4     & 0.571  & 0.629  & 0.029  \\
6     & 0.597  & \textbf{0.700}  & 0.056  & 6     & 0.606 & 0.667  & 0.046  & 6     & 0.630  & 0.692  & 0.034  \\
8     & \textbf{0.627}  & 0.667  & 0.043  & 8     & 0.628  & 0.714  & 0.049  & 8     &\textbf{ 0.700}  & \textbf{0.750}  & 0.024  \\
10    & 0.536  & 0.667  & 0.060  & 10    & \textbf{0.667} & \textbf{0.762}  & 0.052  & 10    & 0.685  & 0.720  & 0.031  \\
12    & 0.519  & 0.636  & 0.066  & 12    & 0.610 & 0.696  & 0.034  & 12    & 0.647  & 0.714  & 0.033  \\
    \midrule
    \midrule
\multicolumn{4}{c|}{W2V2$^\text{ASR}$}      & \multicolumn{4}{c|}{W2V2$^\text{AER}$}      & \multicolumn{4}{c}{WavLM$^\text{AER}$}     \\
\midrule
Block & F1-avg & F1-max & F1-std & Block & F1-avg & F1-max & F1-std & Block & F1-avg & F1-max & F1-std \\
2     & 0.556  & 0.696  & 0.051  & 2     & 0.541  & 0.615  & 0.050  & 2     & 0.537  & 0.600  & 0.022  \\
4     & 0.598  & \textbf{0.700}  & 0.052  & 4     & 0.579  & 0.643  & 0.043  & 4     & 0.627  & 0.690  & 0.027  \\
6     & \textbf{0.639}  & 0.690  & 0.045  & 6     & 0.605  & \textbf{0.737}  & 0.041  & 6     & 0.638  & 0.667  & 0.027  \\
8     & 0.615  & 0.649  & 0.025  & 8     & \textbf{0.640}  & 0.688  & 0.036  & 8     & 0.707  & \textbf{0.786}  & 0.032  \\
10    & 0.558  & 0.645  & 0.040  & 10    & 0.608  & 0.696  & 0.058  & 10    & \textbf{0.720}  & 0.769  & 0.036  \\
12    & 0.531  & 0.615  & 0.054  & 12    & 0.558  & 0.667  & 0.045  & 12    & 0.684  & 0.750  & 0.032 \\
\bottomrule
\end{tabular}
\vspace{-1ex}
\caption{DAIC-WOZ SDD results using the outputs from different intermediate blocks of different foundation models. Highest F1 value in each column shown in bold.}
\label{tab: single SSL}
\end{table*}

\section{Block-wise SSL Representation analysis}
\label{sec: speech ssl}

It has been previously found that the output of different encoder blocks of a speech foundation model contains different levels of information~\cite{pasad2021layer,zheng22f_interspeech}. The block-wise evolution of the representations follows an acoustic-linguistic hierarchy, where the shallowest layers encode acoustic features, followed by the word meaning information, and phonetic and word identities. The analysis of the intermediate block representations can provide insights to better understand the information relevant to SDD. In this section, we perform such an analysis for the first time for SDD. The model structure used for block-wise analysis is shown in Fig.~\ref{fig: updown}(b). Each time output from one intermediate Transformer block from the foundation model was used  for downstream SDD.

\begin{table}[htb]
\centering
\begin{tabular}{c|ccccc}
\toprule
$M^{+}$ & 100    & 200    & 500    & 1000   & 1500   \\
\midrule
F1-avg         & 0.451  & 0.583  & 0.647  & 0.679  & 0.669  \\
F1-max         & 0.640   & 0.700    & 0.714  & 0.762  & 0.727  \\
F1-std         & 0.131 & 0.082 & 0.033 & 0.027 & 0.031 \\
\bottomrule
\end{tabular}
\vspace{-1ex}
\caption{DAIC-WOZ SDD results with increased number of augmented utterances. WavLM$^{\text{PT}}_\text{L12}$ used as input. $M^+$ is the number of sub-dialogues for each positive sample.}
\label{tab: data aug}
\vspace{-2ex}
\end{table}

\subsection{Effect of data augmentation}
The effect of data augmentation was first investigated using the output of the last (12th) Transformer block of the pre-trained WavLM model (WavLM$^{\text{PT}}_\text{12}$). Augmenting data trades off between generating more data and matching the true data distribution. As shown in Table~\ref{tab: data aug}, the F1 score increases and standard deviation decreases as the number of sub-dialogues for each positive sample $M^+$ increases up until 1000, then F1 decreases and standard deviation increases. The model runs the risk of overfitting the training data if each original sequence is replicated too many times. $M^{+}=500$ is used in following experiments, weighing performance and training time.

\subsection{Pre-trained SSL representations}
\label{ssec:ptssl}
The parameters of the three pre-trained foundation models (W2V2$^\text{PT}$, HuBERT$^\text{PT}$, WavLM$^\text{PT}$) were frozen and the SDD results using different intermediate blocks of the models are shown in Table~\ref{tab: single SSL}. F1-avg of the intermediate blocks of three models are plotted in Fig.~\ref{fig: trend}(a). For all three models, F1 first improves as the layer number increases and then F1 decreases. Overall WavLM$^\text{PT}$ produces a F1 score higher than the other two models. Features extracted from the 10th-block give the highest F1 for HuBERT$^\text{PT}$ while features extracted from the 8th-block have the overall best performance for W2V2$^\text{PT}$ and WavLM$^\text{PT}$. It has been found~\cite{pasad2021layer} that the first a few W2V2 Transformer blocks show increased similarity with Mel filter bank (FBank) features, indicating that shallow layers encode acoustic information much like FBank. Word meaning information is mainly encoded in middle blocks, especially around the 8th-block~\cite{pasad2021layer}. Hence it can be inferred that features contain word meaning information are useful for SDD.

\subsection{ASR and AER fine-tuned representations}
\label{sec:ft ssl}
This section investigates how fine-tuning changes the findings in Section~\ref{ssec:ptssl}. It has been implied in Section~\ref{ssec:ptssl} that intermediate layer containing information correlated with word meaning is effective to SDD. And it has been found~\cite{wu2022climate} that emotion information is also useful to SDD. Thus, our foundation models are fine-tuned
based on ASR and AER tasks. Three fine-tuned systems are investigated in this paper: W2V2 base model fine-tuned for ASR on the 960 hours of Librispeech (W2V2$^\text{ASR}$)\footnote{Available at https://huggingface.co/facebook/wav2vec2-base-960h}, W2V2 base model fine-tuned on 110  
\begin{figure}[H]
    \centering
    \begin{minipage}[b]{\linewidth}
    \centerline{\includegraphics[width=0.8\linewidth]{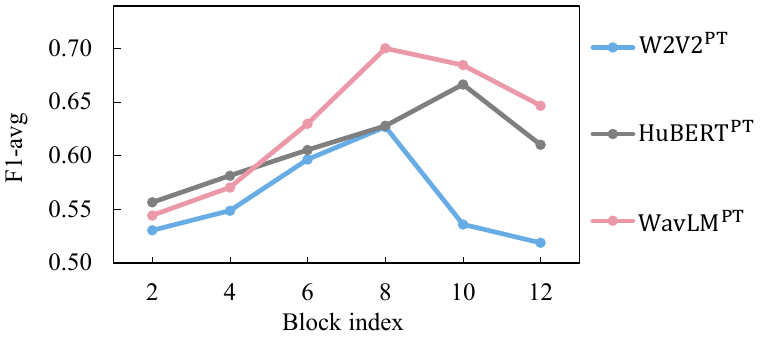}}
    \centerline{(a) Pre-trained foundation models.}
    \vspace{1.5ex}
    \centerline{\includegraphics[width=0.8\linewidth]{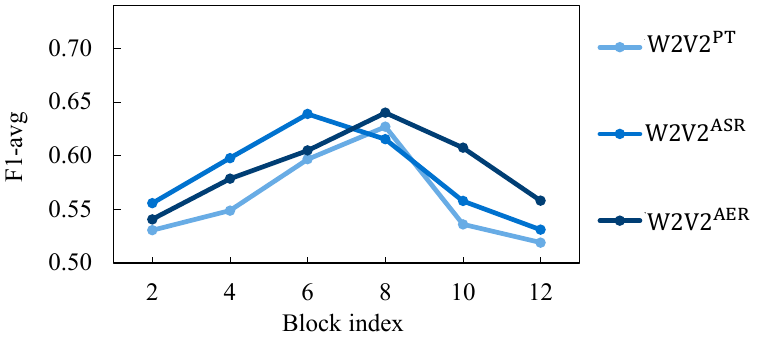}}
    \centerline{(b) W2V2 fine-tuned on ASR and AER.}
    \vspace{1.5ex}
    \centerline{\includegraphics[width=0.8\linewidth]{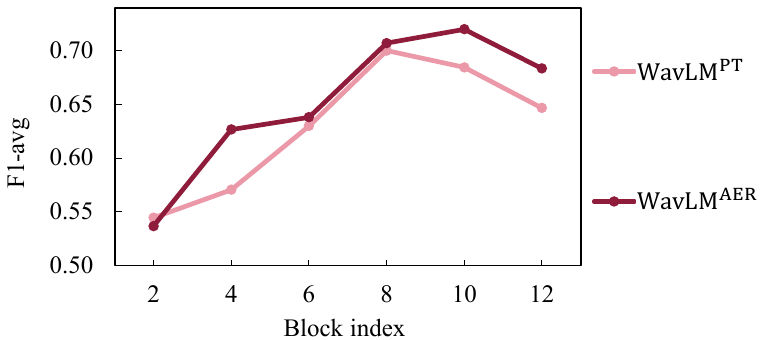}}
    \centerline{(c) WavLM fine-tuned on AER.}
    \end{minipage}
    \caption{Trends of DIAC-WOZ F1-avg values at different blocks for the foundation models.}
    \label{fig: trend}
    \vspace{-2ex}
\end{figure}
\noindent hours of MSP-Podcast dataset~\cite{MSP-podcast} for AER by adding two extra FC layers at the end (W2V2$^\text{AER}$), and WavLM base model fine-tuned in the same way as W2V2$^\text{AER}$ for AER (WavLM$^\text{AER}$). The concordance correlation coefficients for valence, activation, and dominance are 0.418, 0.658, 0.562 for W2V2$^\text{AER}$, and 0.445, 0.667, 0.597 for WavLM$^\text{AER}$. The model parameters were frozen after fine-tuning.

SDD results of the fine-tuned models are shown in the bottom half of Table~\ref{tab: single SSL}. 
Comparing the results of W2V2$^\text{ASR}$ and W2V2$^\text{AER}$ with W2V2$^\text{PT}$ , as shown in Fig.~\ref{fig: trend}(b), the peak of W2V2$^\text{ASR}$ is more towards earlier blocks while the peak of W2V2$^\text{AER}$ is towards later blocks. As shown by Fig.~\ref{fig: trend}(c), the performance of WavLM$^\text{AER}$ also improves over WavLM$^\text{PT}$ on later layers. The fine-tuned foundation models presumably learn more task-specific information. For a W2V2 model fine-tuned with character-level connectionist temporal classification loss~\cite{graves2006connectionist}, the output of the last few layers are more directly related to the word identities. Fine-tuning the foundation model for AER improves the overall performance, indicating that emotion and depression share some para-linguistic indicators encoded by the fine-tuned  models.

\section{The use of ASR transcriptions}
\label{sec: text ssl}
It has been shown that text information is effective for SDD~\cite{williamson2016detecting,dinkel2019text}. However, reference transcriptions are usually not available in practice. This section uses an ASR system to transcribe the depression detection interview and investigates the performance of using erroneous transcriptions in SDD. Transcriptions were obtained from the final output of the W2V2$^\text{ASR}$ model which has a word error rate (WER) of 3.4\% on LibriSpeech test-clean and 8.6\% on test-other while 40.9\% on DAIC-WOZ. The ASR and reference transcripts were encoded by a text foundation model, the RoBERTa base model\footnote{Available at https://huggingface.co/roberta-base}~\cite{liu2019roberta}, and fed into the depression detection block. The SDD results with ASR generated hypotheses and reference transcriptions are compared in Table~\ref{tab: text} (RoBERTa$^\text{Hyp}$, RoBERTa$^\text{Ref}$). Replacing the reference transcriptions with ASR generated hypotheses leads to a decrease of 0.36 in average F1 score and also  a larger standard deviation.

\begin{table}[t]
\centering
\begin{tabular}{cccc}
\toprule
System & F1-avg    & F1-max   & F1-std    \\
\midrule
RoBERTa$^\text{Hyp}$ &0.599 & 0.667 & 0.042 \\
RoBERTa$^\text{Ref}$ &0.635 & 0.667 & 0.029  \\
\midrule
\cat{}$\{$RoBERTa$^\text{Hyp}$,W2V2$^{\text{ASR}}_{6}\}$ & \textbf{0.648} & \textbf{0.714} & 0.028 \\
\bottomrule
\end{tabular}
\vspace{-1ex}
\caption{Comparison of using reference and ASR transcriptions for SDD, where \cat{}$\{\cdot,\cdot\}$ refers to a concatenation.}
\label{tab: text}
\vspace{-2ex}
\end{table}

\begin{table}[b]
\centering
\vspace{-2ex}
\begin{tabular}{cccc}
\toprule
System & F1-avg    & F1-max   & F1-std    \\
\midrule
RoBERTa$^\text{Hyp}$&0.599 & 0.667 & 0.042 \\
WavLM$^{\text{PT}}_{8}$ & {0.700} & {0.750} & {0.024} \\
WavLM$^{\text{AER}}_{10}$ & {0.720} & 0.769 & 0.036 \\
\midrule
\cat{}$\{$WavLM$^{\text{PT}}_{8}$, RoBERTa$^\text{Hyp}\}$ & 0.725 & 0.759 & 0.021 \\
\cat{}$\{$WavLM$^{\text{AER}}_{10}$, RoBERTa$^\text{Hyp}\}$ & 0.756 & 0.800 & 0.023  \\
\bottomrule
\end{tabular}
\vspace{-1ex}
\caption{Results of combining different speech and text foundation models, where \cat{}$\{\cdot,\cdot\}$ refers to a concatenation. }
\label{tab: concatenation}
\vspace{-1ex}
\end{table}

Utterance-level representations derived from RoBERTa$^\text{Hyp}$ were combined with those derived from the 6th-block representations of the ASR-fine-tined W2V2 model (W2V2$^{\text{ASR}}_{6}$) by concatenation. From Table~\ref{tab: text}, this combination produced better SDD results than using the reference transcriptions alone.

\section{combinations of foundation models}
\label{sec:combination}
This section studies further combinations of SSL representations derived from both speech and text foundation models. Similar to the experiments in Table~\ref{tab: text}, speech SSL representations were combined with the ASR transcriptions by a concatenation, and the results are shown in Table~\ref{tab: concatenation}. Combining speech and ASR-hypothesis-based text representations can improve F1-avg and F1-max as well as reduce F1-std, which improves both SDD classification performance and stability.

Finally we investigated the use of a system ensemble by voting. Two ensembles were investigated: (i) The ensemble of systems based on three speech foundation models: W2V2$^\text{PT}_{6}$, HuBERT$^\text{PT}_{10}$, WavLM$^\text{PT}_{8}$; (ii) The ensemble of systems from three modalities: WavLM$^\text{PT}_{8}$ (audio modality),  \cat{}$\{$WavLM$^\text{AER}_{10}$, RoBERTa$^\text{Hyp}\}$ (text modality), WavLM$^\text{AER}_{10}$ (emotion modality).

The results of using ensembles are shown in Table~\ref{tab: ensemble}.
Reference transcriptions were not used in these ensembles and our best-performing depression detection systems require only the speech input.
Table~\ref{tab: cross compare} cross compares our results with those published in literature. Paper~\cite{ravi22_interspeech} used W2V2 and reported the average result across five models. Reference transcriptions were used by papers~\cite{gong2017topic,al2018detecting,shen2022automatic,wu2022climate}. The comparison shows that the ensemble of foundation models produced competitive performance for depression detection based on speech input only.

\begin{table}[tb]
\centering
\begin{tabular}{c|cc}
\toprule
System &   Ensemble 1 & Ensemble 2 \\
\midrule
W2V2$^\text{PT}_{6}$ & $\surd$ &\\
HuBERT$^\text{PT}_{10}$ & $\surd$ & \\
WavLM$^\text{PT}_{8}$ & $\surd$ & $\surd$\\
WavLM$^\text{AER}_{10}$ & &$\surd$\\
\cat{}$\{$WavLM$^\text{AER}_{10}$, RoBERTa$^\text{Hyp}\}$ & & $\surd$\\
\midrule
F1-avg & 0.800 & 0.829 \\
F1-max & 0.857 & 0.886 \\
\bottomrule
\end{tabular}
\vspace{-1ex}
\caption{Ensemble of foundation models.  \cat{}$\{\cdot,\cdot\}$ refers to a concatenation.}
\label{tab: ensemble}
\end{table}

\begin{table}[tb]
\centering
\begin{tabular}{c|ccccc|c}
\toprule
Paper & 
\cite{ravi22_interspeech} &
\cite{gong2017topic} & 
\cite{al2018detecting} &
\cite{shen2022automatic} &
\cite{wu2022climate} & ours \\
\midrule
F1-avg & 0.69 & - & - & - & - & {0.83} \\ 
F1-max  & - & 0.70  & 0.77 & 0.85  & 0.87& \textbf{0.89} \\
\bottomrule
\end{tabular}
\vspace{-1ex}
\caption{Cross comparison on DAIC-WOZ development subset.}
\label{tab: cross compare}
\vspace{-2ex}
\end{table}

\section{Conclusion}
This paper studies the use of SSL representations in speech-based depression detection. 
Block-wise analysis of the foundation models implies that word meaning information is helpful in SDD. 
Fine-tuning pre-trained speech foundation models for AER improves SDD performance, indicating that some indicators are shared between AER and SDD. SDD performance when using  ASR transcriptions matches that of using reference transcriptions when combined with the hidden representations derived from an ASR-fine-tuned foundation model. The ensemble of speech and text foundation models produced the SOTA F1 score of 0.89 on DAIC-WOZ dataset without using the reference transcriptions.

\FloatBarrier
\bibliographystyle{IEEEbib}
\bibliography{refs}

\begin{thebibliography}{10}

\bibitem{WHO}
{World Health Organization},
\newblock ``Depression,''
\newblock {\em
  \url{https://www.who.int/news-room/fact-sheets/detail/depression}}, Accessed:
  August 15, 2022.

\bibitem{cummins2015review}
Nicholas Cummins, Stefan Scherer, Jarek Krajewski, Sebastian Schnieder, Julien
  Epps, and Thomas~F Quatieri,
\newblock ``A review of depression and suicide risk assessment using speech
  analysis,''
\newblock {\em Speech Communication}, vol. 71, pp. 10--49, 2015.

\bibitem{moore2007critical}
Elliot Moore~II, Mark~A Clements, John~W Peifer, and Lydia Weisser,
\newblock ``Critical analysis of the impact of glottal features in the
  classification of clinical depression in speech,''
\newblock {\em {IEEE Transactions on Biomedical Engineering}}, vol. 55, no. 1,
  pp. 96--107, 2007.

\bibitem{low2010detection}
Lu-Shih~Alex Low, Namunu~C Maddage, Margaret Lech, Lisa~B Sheeber, and
  Nicholas~B Allen,
\newblock ``Detection of clinical depression in adolescents’ speech during
  family interactions,''
\newblock {\em {IEEE Transactions on Biomedical Engineering}}, vol. 58, no. 3,
  pp. 574--586, 2010.

\bibitem{ooi2012multichannel}
Kuan Ee~Brian Ooi, Margaret Lech, and Nicholas~B Allen,
\newblock ``Multichannel weighted speech classification system for prediction
  of major depression in adolescents,''
\newblock {\em {IEEE Transactions on Biomedical Engineering}}, vol. 60, no. 2,
  pp. 497--506, 2012.

\bibitem{williamson2016detecting}
James~R Williamson, Elizabeth Godoy, Miriam Cha, Adrianne Schwarzentruber,
  Pooya Khorrami, Youngjune Gwon, Hsiang-Tsung Kung, Charlie Dagli, and
  Thomas~F Quatieri,
\newblock ``Detecting depression using vocal, facial and semantic communication
  cues,''
\newblock in {\em Proc. ACM MM}, Amsterdam, 2016.

\bibitem{al2018detecting}
Tuka Al~Hanai, Mohammad~M Ghassemi, and James~R Glass,
\newblock ``Detecting depression with audio/text sequence modeling of
  interviews.,''
\newblock in {\em Proc. Interspeech}, Hyderabad, 2018.

\bibitem{bommasani2021opportunities}
Rishi Bommasani, Drew~A Hudson, Ehsan Adeli, Russ Altman, Simran Arora, Sydney
  von Arx, Michael~S. Bernstein, Jeannette Bohg, Antoine Bosselut, Emma
  Brunskill, et~al.,
\newblock ``On the opportunities and risks of foundation models,''
\newblock {\em arXiv preprint arXiv:2108.07258}, 2021.

\bibitem{yang2021superb}
Shu-wen Yang, Po-Han Chi, Yung-Sung Chuang, Cheng-I~Jeff Lai, Kushal Lakhotia,
  Yist~Y Lin, Andy~T Liu, Jiatong Shi, Xuankai Chang, Guan-Ting Lin, et~al.,
\newblock ``{SUPERB}: Speech processing universal performance benchmark,''
\newblock {\em arXiv preprint arXiv:2105.01051}, 2021.

\bibitem{baevski2020wav2vec}
Alexei Baevski, Yuhao Zhou, Abdelrahman Mohamed, and Michael Auli,
\newblock ``{Wav2Vec} 2.0: A framework for self-supervised learning of speech
  representations,''
\newblock in {\em Proc. NeurIPS}, Conference held virtually, 2020.

\bibitem{hsu2021hubert}
Wei-Ning Hsu, Benjamin Bolte, Yao-Hung~Hubert Tsai, Kushal Lakhotia, Ruslan
  Salakhutdinov, and Abdelrahman Mohamed,
\newblock ``{HuBERT}: {S}elf-supervised speech representation learning by
  masked prediction of hidden units,''
\newblock {\em IEEE {T}ransactions on Audio, Speech, and Language Processing},
  vol. 29, pp. 3451--3460, 2021.

\bibitem{chen2022wavlm}
Sanyuan Chen, Chengyi Wang, Zhengyang Chen, Yu~Wu, Shujie Liu, Zhuo Chen, Jinyu
  Li, Naoyuki Kanda, Takuya Yoshioka, Xiong Xiao, et~al.,
\newblock ``Wav{LM}: Large-scale self-supervised pre-training for full stack
  speech processing,''
\newblock {\em IEEE Journal of Selected Topics in Signal Processing}, vol. 16,
  no. 6, pp. 1505--1518, 2022.

\bibitem{zhang2022bigssl}
Yu~Zhang, Daniel~S Park, Wei Han, James Qin, Anmol Gulati, Joel Shor, Aren
  Jansen, Yuanzhong Xu, Yanping Huang, Shibo Wang, et~al.,
\newblock ``{BigSSL}: Exploring the frontier of large-scale semi-supervised
  learning for automatic speech recognition,''
\newblock {\em IEEE Journal of Selected Topics in Signal Processing}, vol. 16,
  no. 6, pp. 1519--1532, 2022.

\bibitem{morais2022speech}
Edmilson Morais, Ron Hoory, Weizhong Zhu, Itai Gat, Matheus Damasceno, and
  Hagai Aronowitz,
\newblock ``Speech emotion recognition using self-supervised features,''
\newblock in {\em Proc. ICASSP}, Singapore, 2022.

\bibitem{liu2019roberta}
Yinhan Liu, Myle Ott, Naman Goyal, Jingfei Du, Mandar Joshi, Danqi Chen, Omer
  Levy, Mike Lewis, Luke Zettlemoyer, and Veselin Stoyanov,
\newblock ``{RoBERTa}: A robustly optimized bert pretraining approach,''
\newblock {\em arXiv preprint arXiv:1907.11692}, 2019.

\bibitem{DAIC-WOZ}
D.~DeVault, R.~Artstein, G.~Benn, T.~Dey, E.~Fast, A.~Gainer, K.~Georgila,
  J.~Gratch, A.~Hartholt, M.~Lhommet, G.~Lucas, S.~Marsella, F.~Morbini,
  A.~Nazarian, S.~Scherer, G.~Stratou, A.~Suri, D.~Traum, R.~Wood, Y.~Xu,
  A.~Rizzo, and L.~P. Morency,
\newblock ``{SimSensei Kiosk}: A virtual human interviewer for healthcare
  decision support,''
\newblock in {\em Proc. AAMAS}, Paris, 2014.

\bibitem{shen2022automatic}
Ying Shen, Huiyu Yang, and Lin Lin,
\newblock ``Automatic depression detection: An emotional audio-textual corpus
  and a gru/bilstm-based model,''
\newblock in {\em Proc. ICASSP}, Singapore, 2022.

\bibitem{gong2017topic}
Yuan Gong and Christian Poellabauer,
\newblock ``Topic modeling based multi-modal depression detection,''
\newblock in {\em Proc. ACM MM}, Mountain View, 2017.

\bibitem{ravi22_interspeech}
Vijay Ravi, Jinhan Wang, Jonathan Flint, and Abeer Alwan,
\newblock ``{A Step Towards Preserving Speakers’ Identity While Detecting
  Depression Via Speaker Disentanglement},''
\newblock in {\em Proc. Interspeech}, Incheon, 2022.

\bibitem{wu2022climate}
Wen Wu, Mengyue Wu, and Kai Yu,
\newblock ``Climate and weather: Inspecting depression detection via emotion
  recognition,''
\newblock in {\em Proc. ICASSP}, Singapore, 2022.

\bibitem{pasad2021layer}
Ankita Pasad, Ju-Chieh Chou, and Karen Livescu,
\newblock ``Layer-wise analysis of a self-supervised speech representation
  model,''
\newblock in {\em Proc. ASRU}, Cartagena, 2021.

\bibitem{zheng22f_interspeech}
Xianrui Zheng, Chao Zhang, and Phil Woodland,
\newblock ``{Tandem Multitask Training of Speaker Diarisation and Speech
  Recognition for Meeting Transcription},''
\newblock in {\em Proc. Interspeech}, Incheon, 2022.

\bibitem{MSP-podcast}
R.~Lotfian and C.~Busso,
\newblock ``Building naturalistic emotionally balanced speech corpus by
  retrieving emotional speech from existing podcast recordings,''
\newblock {\em {IEEE Transactions on Affective Computing}}, vol. 10, no. 4, pp.
  471--483, 2019.

\bibitem{graves2006connectionist}
Alex Graves, Santiago Fern{\'a}ndez, Faustino Gomez, and J{\"u}rgen
  Schmidhuber,
\newblock ``Connectionist temporal classification: labelling unsegmented
  sequence data with recurrent neural networks,''
\newblock in {\em Proc. ICML}, Pittsburgh, 2006.

\bibitem{dinkel2019text}
Heinrich Dinkel, Mengyue Wu, and Kai Yu,
\newblock ``Text-based depression detection on sparse data,''
\newblock {\em arXiv preprint arXiv:1904.05154}, 2019.

\end{thebibliography}

\end{document}